# Intelligibilité multimodale de l'hypertexte érudit : le rôle du documentaliste

## Une nécessaire collaboration pour la documentarisation sérielle dans la chaîne éditoriale scientifique

Gérald Kembellec*, **

*  Cnam, Direction de la Recherche, Laboratoire Dicen-IdF
*292 rue Saint Martin, 75003 Paris*

*gerald.kembellec@cnam.fr*
** Institut historique allemand, département des humanités numériques
*8 rue du Parc Royal, 75003 Paris*
*gkembellec@dhi-paris.fr*

RÉSUMÉ. *Cet article montre la porosité des frontières entre les métiers de l'édition et de la publication en ligne et que le nécessaire renouvellement des métiers de l'info-documentation trouverait une place naturelle face à l'évolution du Web. Nous pensons en particulier à l'accompagnement de la délicate mise en œuvre de la documentarisation sérielle des hypertextes érudits, spécifiquement en contextes scientifique ou culturel. La vocation de cet article est de démontrer que la valorisation hypertexte d'un document de qualité au sein des ramifications du Web ne peut se faire qu'à travers une bonne intelligence entre auteur(s), éditeur et diffuseur, pour à terme rencontrer son public. Il sera également démontré que chaque acteur de cette chaîne auctorio-éditoriale serait gagnant dans le cadre d'un travail de formalisation qualitatif qui aurait une portée diffusionelle forte. Enfin, nous pointerons que ce travail d'intermédiation doit être piloté par un acteur de l'info-communication, pour rendre le texte intelligible aux humains comme aux machines. Cet acte médiateur est ici désigné sous le terme de documentarisation sérielle.*

ABSTRACT. *This article shows that the boundaries between the editing and online publishing professions are losing their strength. In this context it would only make sense that the way hypertexts are documented be renewed, especially facing of the Web's evolution. We are thinking in particular of the trickier scholar hypertexts documentation process - specifically in scientific or cultural contexts. The purpose of this article is to demonstrate that, considering the numerous branches of the Web, the hypertext valorisation of a document of quality can only be done through a proper dialogue between authors, editors, and broadcasters. It would satisfy the readership as they could reach the appropriate information. It will also be shown that each actor in this auctorial-editorial process would be a gainer. Indeed, a qualitative formalisation work would be coupled with a strong broadcasting scope. Finally, we will point out that this work of mediating must be led by an actor of information-communication, to make the text understandable to both humans and machines. This mediative act is designated here under the term of serial documentarisation.*

**1. Introduction**

Le monde numérique tel que nous le connaissons est issu d'un système complexe qui s'est imposé dans l'industrie, tout en développant une culture propre (Doueihi, 2013). Il se situe à l'intersection des sciences computationnelles techniques et de celles de l'information, projetant le calculatoire sur l'information, créant et diffusant ainsi de la connaissance (Verlaet, 2015). Verlaet note une rupture au XX$^e$ siècle dans la logique de codage et d'usage de l'information à l'échelle mondiale. Elle rappelle, interprétant Bourdieu, que notre monde « a été le théâtre — pour le moins prolifique — de la création et du développement des technologies numériques » (*ibid*).

Les textes érudits, qu'ils soient littéraires, scientifiques ou culturels subissent de nombreuses mutations de fond et de forme depuis que la question même de l'érudition fait sens dans l'opinion lettrée. Ces évolutions sont étudiées en épistémologie, en histoire des sciences, en sociologie des médias, en histoire du livre de même qu'en sciences de l'information et de la communication. Les modes d'écritures et d'interactions érudites se succèdent, se croisent et se complètent depuis des millénaires dans une sociabilité du texte qui, sans forcément se renouveler totalement à travers les époques, n'a cessé d'évoluer pour aboutir à son incarnation hypertexte (Blair, 2020; Grafton, 2015; Kembellec, 2021; Le Deuff, 2015). Dans cet article, nous proposons une analyse critique des modèles de l'édition numérique érudite actuelle dans sa forme hypertexte. Nous expliquons que la spécificité même de ce support renégocie la collaboration entre l'auteur, l'éditeur, le diffuseur et bien sûr le lecteur. Les récentes évolutions sémantiques du maillage hypertexte, les strates de codifications, la multiplicité des publics et agents d'interprétation transforment la relation entre les acteurs de l'écriture, de l'information, de la publication et de la diffusion du texte érudit en ligne.

Nous affirmons et aspirons à démontrer que la codification multimodale de l'hypertexte érudit est un prérequis à l'efficacité diffusionnelle du message véhiculé. En effet, selon l'intelligibilité du texte proposé, celui-ci sera compréhensible par tous les publics, le lectorat érudit, les publics empêchés appareillés et même les acteurs non humains comme les agents d'indexation. Nous parlons ici de l'intelligibilité du texte lui-même, mais aussi de la manière dont son inscription numérique le rend intelligible à travers les strates de l'hypertexte. Pour maximiser la portée diffusionnelle du document numérique, il faut anticiper sa réception par les lectorats et penser leurs accompagnements.

## 2. Le document scientifique en ligne, un objet frontière

Nous plaçons notre cadre théorique dans les modèles formels d'hypertextes qui se croisent au sein des architectures hypermédias comme catalyseurs épistémiques dans la complexe tâche d'écri-lire et de publier des documents scientifiques en ligne (Crozat et al., 2011; Saleh & Hachour, 2012; Verlaet et al., 2013). Plus spécifiquement, nous centrons nos réflexions sur les usages et besoins des différents acteurs qui interagissent supposément autour d'un hyperdocument scientifique. Les points qui découlent de cette analyse mettent en tension les fondements mêmes de l'hypertexte. Nous pensons ici à la navigation, mais aussi à l'accessibilité multimodale des contenus par la description. Ce point croise la question des autorités de description du Web en contexte scientifique, les normes, formats et standards qui sont les prérequis de tout système communiquant.

Cette approche critique des hypertextes scientifiques, appliquée au domaine des sciences de l'information et de la communication convoque bien sûr la théorie opérationnelle de l'écriture en contexte de réseau numérique hypertexte. Elle fait aussi appel à celle de la mise en œuvre plus spécifique de la documentation et de la documentarisation, dans une optique auctoriale et éditoriale telle que présentée depuis des années collectivement dans Pédauque et plus individuellement avec Salaün, Vitalli-Rossati ou encore Zacklad… (J. M Salaün, 2007; Pédauque, 2003, 2006; Vitali Rosati, 2016; Zacklad, 2018, 2019).

Ces analyses de la (re)documentarisation posent d'autres questions que celles de la mise en œuvre d'une présentation de fragments informationnels dans le réseau hypertexte sériel que peut constituer l'écosystème d'une revue en ligne. Ces questions, plus vastes touchent à des problématiques liées à la sociologie des médias, à l'éthique du numérique et peuvent même trouver des convergences avec la mercatique numérique pour les questions de la visibilité, de soi et de la revue.

Faisant fi d'un débat postural que l'on peut espérer clôt sur la place de la médiologie en regard des SIC et plus largement des SHS, les pistes évoquées précédemment nous amènent à croiser la médiologie. Cela permet de comprendre, en l'état actuel et en anticipation, la manière dont le régime de la graphosphère se projette dans celui l'hypersphère au sein de la médiation scientifique en ligne (Debray, 1991; Merzeau, 1998; S. N., 1998). L'hypertexte scientifique est donc un objet frontière, à l'intersection de plusieurs disciplines, dont la rencontre nécessite des aptitudes techniques à la médiation sur plusieurs niveaux de lecture. Commençons par analyser les enjeux propres au document scientifique, plus spécifiquement à sa mise en média au sein d'un réseau hypertexte.

## 3. Le document scientifique en ligne et ses enjeux

Si l'on se place dans une perspective humaniste et littéraire avec les concepts de lecture « distante », par exemple avec l'assistance de dispositifs de détection formelle,

« attentive » en écrit d'écran ou même « citationnelle » avec un logiciel de gestion de références bibliographiques, les attentes en termes de préparation du texte ne sont pas les mêmes (Moretti, 2000, 2003; Souchier, 1996). Pourtant, malgré ces enjeux différents, le point d'accès reste unique : un document numérique en ligne chargé et analysé par une machine qui en propose une restitution selon plusieurs modalités. Comment combiner les réponses à toutes ces exigences formelles quand l'accès à l'hypertexte est unique sur le Web ? Comment pallier l'inadéquation entre les modèles diffusionnels éditoriaux et les besoins des lectorats ? L'absence de réponse à ces questions peut être préjudiciable à tous les acteurs gravitant autour de l'hypertexte érudit : l'auteur risque d'être moins lu (et moins cité), l'éditeur perdra également en lectorat et donc en métriques d'évaluation quantitative pour son ouvrage ou sa revue. Quant au diffuseur, pour ne pas être en reste, il risque de voir son modèle économique, quel qu'il soit, péricliter. Mais, hors de ces questions égotiques ou mercantiles, l'auteur d'un texte de qualité risque d'être privé de son lectorat et réciproquement. Dans un contexte d'abondance informationnelle, le lecteur impatient pourrait, dans l'absolu, faire l'impasse sur un texte pertinent pour des questions de formalismes provoquant une inadéquation avec ses pratiques.

### 3.1. *Les besoins (supposés) des chercheurs en termes de médiation*

Nous avions, dans des articles précédents, synthétisé les travaux exploratoires qui ont mis en exergue les besoins érudits des chercheurs en sciences humaines et sociales : des « primitives conceptuelles érudites ». Ces primitives viennent trouver leur matérialité dans un périmètre de recherche, d'accès et de partage, pour la lecture, l'écriture et la critique. Ces problématiques peuvent être présentées de diverses manières, mais restent peu ou prou les mêmes, quel que soit l'angle d'observation (Unsworth, 2000; d'après ; Hennicke et al., 2017; Palmer et al., 2009). Pour aller à l'essentiel, le chercheur doit échanger de manière bidirectionnelle, écrire, lire, être lu et discuté. Pour cela, il doit accéder de manière efficace à des contenus personnalisés directement en lien avec ses besoins et disposer d'outils ergonomiques qui rendront sa lecture « équipée » (Bigot, 2018). Bien évidemment, il doit veiller à proposer ses écrits dans les mêmes modalités. On l'a compris, la qualité de la médiation est un enjeu fort pour l'auteur et le lecteur qui se trouvent aux deux extrémités de la chaîne éditoriale scientifique : il faut être « lisible » et être capable de trouver de l'information dans un immense corpus. De plus, dans une perspective plus individuelle, Latour a mis en évidence que les chercheurs étant « humains », ils ont un besoin plus ou moins accentué selon les égos de reconnaissance et de valorisation de leurs travaux (Latour, 2001).

### 4. Quelques exemples concrets de revues : des points de vue différents

### 4.1. *Processus et approches d'évaluation*

Ce que le document numérique a de particulier, c'est qu'il n'est pas — contrairement à son ancêtre le codex — autoporteur de sa propre matérialité. Il est donc nécessairement médié sur un dispositif de lecture qui n'est pas, loin de là, uniforme. Dans le cadre de cet article, la présentation d'un hypertexte scientifique en ligne peut prendre un vaste panel de formes. Les façons d'inscrire le document dans un média et de le documentariser vont bien sûr avoir un impact sur les usages qui peuvent en être faits. Il faut donc mettre en question la médiation dans le cadre médiologique des « deux corps du médium », de manière technique et organisationnelle. Cette vision est, si ce n'est complètement holistique, au moins techno-organisationnelle : il faut, dans un artefact de médiation, selon Merzeau (2011, p. 62), inscrire, organiser, réguler et anticiper. Expliquons cet emprunt à la médiologie : la « matière organisée » ou plus simplement la technique de médiation ne peut entrer en résonance avec le lectorat que dans l'« organisation matérialisée » d'un consensus social. En d'autres termes, la technique à elle seule ne résoudra pas des questions sans l'adhésion sociale. La question technico-politique distingue sans opposer les intelligences « collective » et « connective ». Pour interconnecter et augmenter les intelligences humaines, il faut un ou des média(s) qui rassemble(nt) des communautés d'intérêts et de pratiques (Lévy, 1994; Kerckhove, 2000). Lorsqu'il n'existe pas de communauté, ni de pratiques, les nouveaux entrants peuvent proposer librement et les usagers disposer. Dans le cas spécifique de la publication et de l'édition scientifique, tout préexiste : la communauté et ses besoins (Unsworth, 2000), ses pratiques et des acteurs traditionnels issus de la graphosphère et de l'hypersphère. Si les éditeurs et auteurs classiques proposent encore peu de médiation croisée avec des hypermédias[1], ceux de l'hypermédiation ne peuvent en faire l'économie, car c'est l'essence même de ce médium. Cependant, toutes les pratiques ne sont pas équivalentes et n'offrent pas les mêmes conditions d'usage. Prenons quelques exemples qui vont faire apparaître une gradation des modalités de lecture et d'indexation.

### 4.2 *La simple transposition Web du régime de la graphosphère*

Le premier exemple est un travail *a minima* : la simple mise en œuvre du texte dans un document à vocation d'écrit d'écran ou d'imprimé. À une URL spécifique, l'éditeur met à disposition du lectorat des pages HTML ou un fichier PDF. Cette méthode est la stricte transposition d'un *rotulus* dans le cadre HTML ou d'un codex dans le cadre PDF. Il s'agit plus là de la transposition numérique d'un document physique que d'un usage de l'hypertexte, même s'il existe éventuellement des hyperliens vers des sources. Nous n'illustrerons pas d'exemple ces pratiques d'un autre âge du Web, mais il est encore courant de retrouver dans la tradition éditoriale scientifique un simple plaquage de la tradition écrite, sans tentative de sortir de la graphosphère. La notice du document n'est pas décrite dans le code, ce dernier n'intègre pas de métadonnées de valorisation de l'article, de l'auteur, de la revue ou de la plateforme. La documentation des contenus est donc là entièrement déléguée à l'intelligence artificielle des moteurs de recherche avec les biais qu'on lui connaît en contexte d'infobésité (O'neil, 2016).

---

[1] Notons cependant que des ouvrages scientifiques physiques offrant des extensions documentaires dans l'hypershère existent y compris en SHS, voir le billet de Christian Jacob : https://lieuxdesavoir.hypotheses.org/1104. Par exemple, le futur ouvrage historique de Thomas Maissen : *Wie die Jungfrau zum Staat kam. Die Personifikation des Politischen in der Vormoderne, 2023*, prévu dans la collection Pariser Historische Studien chez Heidelberg University Publishing, sera composite entre texte papier et illustrations en ligne, ces dernières étant soigneusement indexées et documentées.

**4.3** *La médiation érudite et la documentarisation sérielle dans l'hypersphère*

Ensuite, pour correspondre à plus d'usages, d'autres revues ou plateformes proposent, outre le texte, la valorisation des métadonnées pour la pratique érudite. Des outils spécifiques de lecture équipée[2] vont, grâce à ces méthodes de préparation de l'hypertexte, proposer d'enregistrer la notice de l'article et celles des articles cités. En plus de cela, il est parfois prévu des modalités de lectures alternatives. Nous n'évoquons pas la simple dé-linéarisation et les allers-retours dans la lecture offerte par l'hypertexte, mais la possibilité d'avoir une autre expérience, plus immersive grâce au formalisme du langage HTML. Ces techniques liées au renouveau sémantique et structurel du HTML, permettent de valoriser le contenu par une forte liaison fond/forme qui fait du sens dans la structure même de l'article ou du document scientifique en général[3].

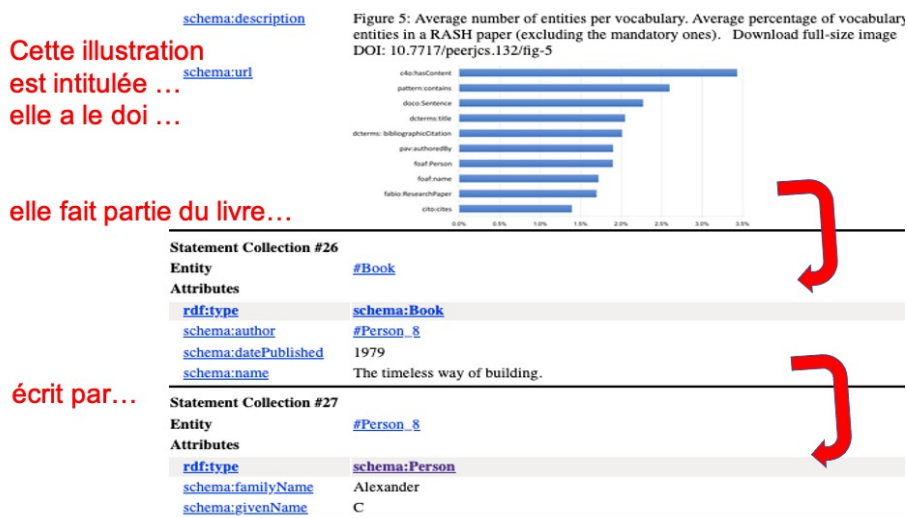

**Figure 1.** *Les langages RDFa et Microformats désambiguïsent acteurs et organisations (vue en lecture équipée avec plug-in de navigation sémantique).*

Ce cadre propose également un contexte matériel rendant possible l'intégration de la sémantique avancée offrant une lecture associée à un péritexte scientifique augmenté de lectures connexes, d'inclusion de contenus issus de Wikidata ou d'autres sources de connaissances fortement structurées. Cela offre les conditions nécessaires à la désambiguïsation de concepts ou de personnes à l'aide de schémas et d'autorités (voir Figure 1) et va permettre de réaliser une lecture distante sur de gros corpus, à la fois pour les moteurs de recherche, le *distant reading* et la lecture érudite augmentée (Goyet, 2017; Kembellec, 2016, 2020; Sire, 2018). Cette pratique offre la possibilité d'un texte fortement structuré, d'une indexation automatique fine mais aussi d'une citabilité augmentée pour des fragments d'articles ou des illustrations sans pour autant apporter d'ajout de connaissance dans le texte.

---

[2] Voir (Bigot, 2018), précédemment cité.
[3] A ce propos voir l'article sur le *Reshearch Article in Simplified HTML* décrit par Peroni et al qui offre un état de l'art des formalismes d'articles scientifiques en ligne et leur compatibilité sémantique (Peroni et al., 2017).

L'usage de ces bonnes pratiques a un effet corollaire non négligeable d'un point de vue diffusionnel. Le consortium des moteurs de recherche valorise et exploite ces contenus considérés comme « riches » au sein d'un réseau d'informations inter reliées : le *Knowledge Graph*. Dans l'exemple présenté dans la capture d'écran en Figure 2, les informations prosopographiques (triviales pour l'exemple) vont être affichées à l'écran, mais aussi encodées par des schémas d'autorité afin que la vie d'Apollinaire soit compréhensible par un automate selon les règles du *semantic publishing* (Shotton, 2009). Ainsi, un outil d'analyse — même rudimentaire — pourra déduire ses lieux et dates de mort ou les pseudonymes qu'il a utilisés. Cette valorisation participe grandement à l'indexation et donc à la visibilité d'un article, de son auteur et également de son contexte de publication : revue, éditeur ou plateforme. Outre le fait de participer à la lisibilité des sujets de l'auteur dans l'hypersphère très encombrée, elle répond à plusieurs des besoins de chercheurs en SHS tels que ceux sociologiques mis en avant par Latour et scientifiques selon Unsworth précédemment cités : la diffusion des idées, leur correcte indexation et attribution, et donc la création de reconnaissance et notoriété par les métriques associées à la recherche.

**Figure 2.** *Création de connaissance à partir des informations en médiation sérielle*

À l'opposé des bénéfices sémantiques observables en *distant reading*, pour celles et ceux qui souhaiteraient une lecture attentive et linéaire, il peut être substitué au mode « sémantique » enrichi, un mode simplifié altgernatif. En effet, la structuration et la documentation d'un hypertexte vont également servir à offrir un mode *close reading* qui va aider à la présentation et favoriser la lecture traditionnelle, mettant la focale sur le texte. Pour aller plus loin, ces mêmes principes vont favoriser la lecture pour les personnes empêchées (voir capture en Figure 3) avec les descriptions structurées des visuels pour la lecture automatique de page Web. Le cadre dispositif des nouveaux navigateurs Web propose en ce sens des fonctionnalités de lecture audio pour personnes empêchées et de lecture immersive pour le *close reading* (Cambre et al., 2021; Williams et al., 2020).

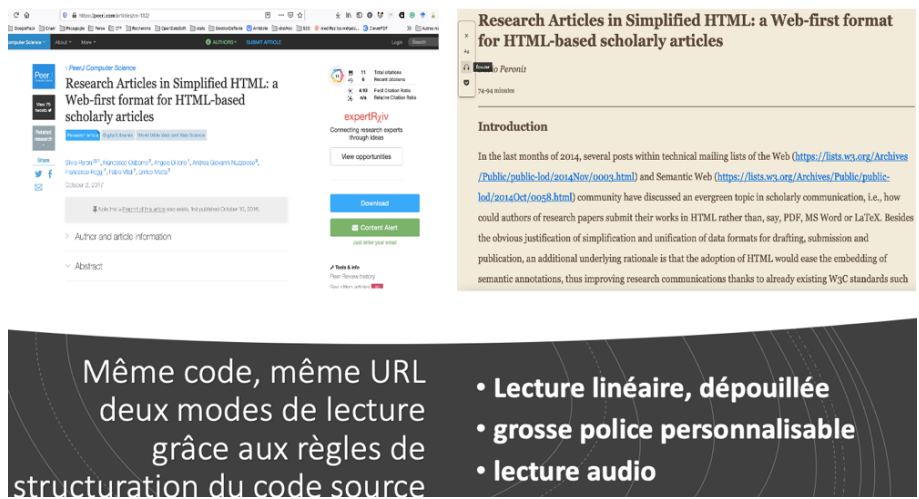

**Figure 3.** *Un même article structuré pour autoriser à la fois la lecture « classique », le rendu « immersif » et la lecture audio.*

### 5. Les qualités convoquées pour l'hypermédiation d'un document scientifique

Nous pensons que l'accompagnement de la délicate mise en œuvre des hypertextes érudits spécifiquement en contextes scientifique ou culturel pourrait être le fruit d'un dialogue entre l'auteur, pour appuyer son intentionnalité, l'éditeur qui souhaite une cohésion structuro-conceptuelle dans sa revue et le diffuseur qui a besoin d'une maximisation de la visibilité des contenus qu'il met à disposition. Cependant, la médiation de l'article scientifique dans une hypersphère concurrentielle ne peut pas être improvisée. On l'a compris, trouver l'équilibre de la médiation et de la documentarisation sérielle a des enjeux en termes d'ergonomie et d'indexabilité.

> Ce que nous entendons par le terme de **documentarisation sérielle** est la mise en œuvre de techniques de documentation de fragments informationnels dans un hypertexte en série (comme une revue ou une plateforme de revues) avec des médiations informatiques à vocation de qualification érudite et diffusionelle.

### 4.1. *Des qualités d'indexation éditoriale Web : segmenter et valoriser*

Il existe bien sûr des architextes de type « gestionnaires de contenus » qui ambitionnent de proposer un système éditorial partiellement automatisé, mais qui ne sont pas encore capables d'offrir une documentarisation fine des fragments informationnels. Déléguer la tâche de documentarisation à l'auteur n'est pas une option viable, ce dernier ne pouvant pas systématiquement être mis à contribution. Il semble évident que ces tâches doivent être accompagnées par des tiers spécialisés dans la détection, la sélection et la valorisation de contenus en collaboration avec les auteurs et les éditeurs.

### 4.2. *Qualités techniques : connaissance du formalisme HTML et Web sémantique*

Une fois le travail intellectuel de valorisation effectué, sa mise en œuvre reste à réaliser à l'aide des techniques, schémas et vocabulaires du Web sémantique. On l'a vu, certaines revues et plateformes spécialisées, avec un modèle économique « auteur-payeur », comme *PeerJ* offrent une segmentation et une structuration de qualité. Malgré le coût induit, ces plateformes n'offrent pas ou peu de réelle documentation des contenus. La valorisation en termes de citabilité et de lecture est optimale, très granulaire, mais l'indexation reste faible. Certains architextes à forte valeur ajoutée, comme c'est le cas avec le dispositif Stylo[4] délèguent ce travail aux auteurs. Cela sous-entend que l'auteur soit à la fois responsable et capable d'indexer, de fragmenter et de valoriser ses contenus. Une communauté d'auteurs est en capacité et a la volonté d'effectuer cette tâche comme le montre la réussite de l'expérience. Cela amène à des résultats extrêmement qualitatifs, dans l'optique de répondre aux besoins des chercheurs, à la condition expresse que l'auteur soit (1) en capacité technique de s'approprier *a minima* le code et (2) dispose de la volonté et du temps pour le faire. Nous estimons que cette solution de documentarisation n'est malheureusement pas sérialisable au plus grand nombre sans une aide de spécialistes du code et de la documentation.

L'idéal dans le cadre d'un éditeur-diffuseur qui souhaiterait réaliser une documentarisation sérielle de qualité dans le cadre de sa ou ses revues, serait de disposer d'un spécialiste des codes des milieux de la recherche, qui dispose de fortes capacités à indexer et documenter. Ce collaborateur disposerait également d'une aptitude à intégrer le résultat de sa collaboration avec les auteurs et l'éditeur dans le document final, soit seul soit en lien avec l'équipe technique de l'éditeur. Ce collaborateur précieux existe et c'est le documentaliste tel qu'il est formé depuis une dizaine d'années dans les universités et les écoles spécialisées comme L'École nationale supérieure des sciences de l'information et des bibliothèques ou l'Institut national des techniques de la documentation pour ne parler que de la France.

---

[4] Voir https://stylo.huma-num.fr/

**Discussion**

Les frontières entre les métiers de l'édition et de la publication en ligne deviennent de plus en plus poreuses. Le métier de l'info-documentation subit depuis l'arrivée du Web une supposée « obsolescence » dans ses rôles traditionnels, ce qui s'ancre dans la réalité du marché de l'emploi et provoque une nécessaire évolution du métier. Nous pensons, comme Accart que cette profession trouverait une place naturelle face à l'évolution du Web, les compétences des acteurs évoluant au même rythme que le Web. En effet, qui serait le mieux à même de documenter les articles scientifiques en lien avec les auteurs que les documentalistes eux-mêmes (Accart, 2020; Accart & Réthy, 2015; Libmann, 2014) ? Nous pensons en particulier, en puisant notre argument sur les modèles de qualité dans la littérature sur les relations sociotechniques en termes de documentarisation sérielle et auctoriale (Zacklad, 2018, 2019), qu'il y a un réel impensé dans les revues scientifiques, même celles nativement hypertextes. La question réelle n'est pas si le documentaliste peut trouver sa place dans une revue scientifique, mais qui va financer ce poste clé. Les éditeurs, auteurs et relecteurs de revues scientifiques sont souvent des enseignants-chercheurs à l'agenda serré et au rôle quasi bénévole dans ce cadre. Il faut rappeler, dans un même temps que l'industrie de l'édition et de la distribution de littérature scientifique est un marché qui se porte extrêmement bien[5].

**Conclusion**

Sans aller jusqu'à l'extrémisme d'un Mc Luhan, arguant que le *medium* serait le message, nous détournons légèrement la pensée *documentologique* de Ferraris en affirmant que l'on pourrait également évaluer un document en ligne par sa valeur interactionnelle, son impact dans un monde social. La question se pose alors : « puis-je réaliser facilement mon activité de chercheur, auteur comme lecteur, en utilisant ou produisant un document en ligne sur une plateforme dédiée ? ». Une réponse affirmative, encore rare, sous-tend une organisation transactionnelle dans la procédure de médiation et de documentarisation sérielle. Cette chaîne éditoriale convoque tout à la fois l'intentionnalité de l'auteur et de l'éditeur, mais aussi les contraintes intégratives et diffusionnelles du distributeur. Dans ces conditions, il est évident qu'un métier de documentaliste-intégrateur scientifique Web doive devenir une réalité ancrée dans les pratiques éditoriales. Une étude qualitative à venir doit questionner l'intérêt réel et l'adhésion des acteurs de l'édition scientifique en ligne : auteurs, éditeurs et diffuseurs, mais aussi la viabilité économique d'un modèle qui équilibrerait les bienfaits qualitatifs et communicationnels avec les coûts induits.

---

[5] En 2019, RELX (Reed Elsevier) affichait un chiffre d'affaires de 7,8 milliards de livres pour un résultat d'exploitation de 2 milliards et demi de livres soit 2,8 milliards d'euros. https://www.zonebourse.com/cours/action/RELX-PLC-9590187/fondamentaux/

La vocation de cet article était de démontrer, à travers des cas d'études — ici des revues scientifiques — que la valorisation d'hypertexte de qualité au sein des ramifications du Web ne peut se faire qu'au travers d'une bonne intelligence entre auteurs, éditeur et diffuseur, pour à terme rencontrer son public. Chaque acteur de cette chaîne auctorio-éditoriale serait gagnant dans le cadre d'un travail de formalisation qualitatif ayant une portée diffusionelle forte. Enfin, nous avons pointé que ce travail d'intermédiation doit être piloté par un acteur de l'info-communication, le documentaliste spécialiste de l'intégration scientifique Web pour rendre le texte intelligible aux humains comme aux machines.

**Bibliographie**

Accart, J.-P., Les bibliothécaires à l'heure de la surabondance d'information : Compétences clés et nouveaux rôles. *Bulletin des Bibliothèques de France*, 2020, https://bbf.enssib.fr/consulter/bbf-2020-00-0000-007

Accart, J.-P., & Réthy, M.-P., *Le Métier de Documentaliste*. Éditions du Cercle de la Librairie, 2015, https://doi.org/10.3917/elec.reth.2015.01

Bigot, J.-É., *Instruments, pratiques et enjeux d'une recherche numériquement équipée en sciences humaines et sociales* [PhD Thesis]. Université de Technologie de Compiègne, 2018.

Blair, A. M., *Tant de chose à savoir, comment maîtriser l'information à l'époque moderne.* (B. Krespine, Trad.). Seuil, 2020.

Cambre, J., Williams, A. C., Razi, A., Bicking, I., Wallin, A., Tsai, J., Kulkarni, C., & Kaye, J., *Firefox Voice : An Open and Extensible Voice Assistant Built Upon the Web,* 2021.

Crozat, S., Bachimont, B., Cailleau, I., Bouchardon, S., & Gaillard, L., Éléments pour une théorie opérationnelle de l'écriture numérique. *Document numérique*, *14*(3), 2011, p. 9-33.

Debray, R., *Cours de médiologie générale,* 1991.

Goyet, S., Outils d'écriture du web et industrie du texte. *Réseaux*, *6*, 2017, p. 61-94.

Grafton, A., *La page de l'antiquité à l'ère du numérique. Histoire, usages, esthétiques*. Hazan, 2015, https://www.editions-hazan.fr/livre/la-page-de-lantiquite-lere-du-numerique-histoire-usages-esthetiques-9782754108126

Hennicke, S., Gradmann, S., Thoden, C., Dill, K., Tschumpel, G., Morbindoni, K., & Pichler, A., *Research Report on DH Scholarly Primitives* (Commission européenne ICT-PSP-297274), 2017.

Kembellec, G., Que voit réellement Google de la sémantique des pages web ? *I2D – Information, données & documents*, 53(2), 2016. p. 65.

Kembellec, G., & Bottini, T., Réflexions sur le fragment dans les pratiques scientifiques en ligne : Entre matérialité documentaire et péricope. *20° Colloque International sur le Document Numérique: CiDE. 20*, 2017

Kembellec, G., L'érudition numérique palimpseste. *Hermes, La Revue,* n° 87(2), 2021, p.145-158.

Kerckhove, D. de., *L'intelligence des réseaux*. Odile Jacob, 2000.

Latour, B., *Le métier de chercheur. Regard d'un anthropologue : 2e édition revue et corrigée*. Quae, 2001.

Le Deuff, O., Les humanités digitales précèdent-elle le numérique ? *H2PTM'15: Le numérique à l'ère de l'Internet des objets, de l'hypertexte à l'hyper-objet*, 2015, p. 421-432.

Lévy, P, *L'intelligence collective*, 1994.

Libmann, A.-M., Entretien avec Jean-Philippe Accart: « Les nouveaux documentalistes : Entre survie et valeur ajoutée de service ». *Bases*, n°*319*, 2014, p. 9-11.

Merzeau, L., Ceci ne tuera pas cela. *Les Cahiers de médiologie*, n°2, 1998, p. 27-39.

Merzeau, L., *Pour une médiologie de la mémoire* (HDR, Université de Nanterre-Paris X), 2011.

Moretti, F., Conjectures on world literature. *New left review*, n°1, 2000, p. 54-68.

Moretti, F., More Conjectures. *New left review*, n°20, 2003, p. 73-81.

O'neil, C., *Weapons of math destruction : How big data increases inequality and threatens democracy*. Crown, 2016.

Palmer, C. L., Teffeau, L. C., Pirmann, C. M., *Scholarly information practices in the online environment : Themes from the Literature and Implications for Library Service Development*, Research nº 298733494, 2019, p. 59, OCLC Research. www.oclc.org/programs/publications/reports/2009-02.pdf

Pédauque, R. T, *Document : Forme, signe et médium, les re-formulations du numérique*, 2003.

Pédauque, R. T., *Le document à la lumière du numérique*. C&F Éditions, 2006.

Peroni, S., Osborne, F., Iorio, A. D., Nuzzolese, A. G., Poggi, F., Vitali, F., & Motta, E., Research Articles in Simplified HTML : A Web-first format for HTML-based scholarly articles. *PeerJ Computer Science*, *3*, e132, 2017, https://doi.org/10.7717/peerj-cs.132

S. N., Réponses. *Les cahiers de médiologie*, n°6(2), 1998, p. 285-293, https://www.cairn.info/revue-les-cahiers-de-mediologie-1998-2-page-285.htm

Saleh, I., & Hachour, H., Le numérique comme catalyseur épistémologique. *Revue française des sciences de l'information et de la communication*, n°1(1), 2012, https://doi.org/10.4000/rfsic.168

Salaün, J.-M., La redocumentarisation, un défi pour les sciences de l'information. *Études de communication*. langages, information, médiations, n°30, 2007, p. 13-23. https://doi.org/10.4000/edc.428

Shotton, D., Semantic publishing: the coming revolution in scientific journal publishing. *Learned Publishing*, 2009, vol. 22, n° 2, p. 85-94.

Sire, G., Web sémantique : Les politiques du sens et la rhétorique des données. *Les Enjeux de l'information et de la communication*, 19(2), 2018, p. 147-160.

Souchier, E., L'écrit d'écran, pratiques d'écriture & informatique. *Communication & langages*, 107(1), 1996, p. 105-119.

Unsworth, J., Scholarly primitives : What methods do humanities researchers have in common, and how might our tools reflect this. *Symposium on Humanities Computing: Formal Methods, Experimental Practice. King's College, London*, 13 mai 2000, http://johnunsworth.name/Kings.5-00/primitives.html

Verlaet, L., La deuxième révolution des systèmes d'information : Vers le constructivisme numérique. *Hermes, La Revue*, n° 71(1), 2015, p. 249-254.

Verlaet, L., Gallot, S., & Aguilar, A. G., Le paradigme de la complexité. Apports pour les approches formelles de l'hypertexte. *H2PTM'13, 2013*.

Vitali Rosati, M., Qu'est-ce que l'éditorialisation ? *Sens public,* 2016, http://sens-public.org/articles/1184/

Williams, A. C., Cambre, J., Bicking, I., Wallin, A., Tsai, J., & Kaye, J., Toward Voice-Assisted Browsers : A Preliminary Study with Firefox Voice. *Proceedings of the 2nd Conference on Conversational User Interfaces*, 2020, p. 1-4.

Zacklad, M., Nouvelles tendances en organisation des connaissances. *Études de communication. Langages, information, médiations*, n°50, 2018, https://doi.org/10.4000/edc.4017

Zacklad, M., Le design de l'information : Textualisation, documentarisation, auctorialisation. *Communication langages*, n°199(1), 2019, p. 37-64.